\documentclass{article}
\usepackage{spconf,amsmath,graphicx}
\ninept

\usepackage{booktabs}
\usepackage{multirow}
\usepackage{xcolor}
\usepackage{stfloats}
\usepackage{float}
\usepackage{amsmath}
\usepackage{amsfonts}
\usepackage{amssymb}
\usepackage{blindtext}
\usepackage{cite}
\usepackage[nolist]{acronym}

\DeclareMathOperator*{\argmax}{arg\,max}

\usepackage{pifont}
\newcommand{\cmark}{\ding{51}}%
\newcommand{\xmark}{{\color{white}\ding{55}}}
\newcommand{\figcaption}[2]{
\caption{\textbf{#1} #2}
}
\newcommand{\etal}{\textit{et al}. }
\newcommand{\nparagraph}[1]{\noindent\textbf{#1.  }}

\usepackage{caption}

\title{Lightweight Multi-Branch Network for Person Re-Identification}
\name{Fabian Herzog$^*$ \quad Xunbo Ji\sthanks{$^*$ Equal contribution. Correspondence to: fabian.herzog@tum.de.} \quad Torben Teepe \quad Stefan Hörmann
\quad Johannes Gilg \quad Gerhard Rigoll \thanks{We gratefully acknowledge the financial support from Deutsche Forschungsgemeinschaft (DFG) under grant number RI 658/25-2.}\thanks{© 2021 IEEE. Personal use of this material is permitted. Permission from IEEE must be obtained for all other uses, in any current or future media, including reprinting/republishing this material for advertising or promotional purposes, creating new collective works, for resale or redistribution to servers or lists, or reuse of any copyrighted component of this work in other works.}}
\address{Technical University of Munich}

\usepackage{hyperref}
\begin{document}

\begin{acronym}
\acro{PREID}{Person Re-Identification}
\acro{CE}{Cross-Entropy}
\acro{MS}{Multi-Similarity}
\end{acronym}

\maketitle
\begin{abstract}
Person Re-Identification aims to retrieve person identities from images captured by multiple cameras or the same cameras in different time instances and locations. Because of its importance in many vision applications from surveillance to human-machine interaction, person re-identification methods need to be reliable and fast. While more and more deep architectures are proposed for increasing performance, those methods also increase overall model complexity. This paper proposes a lightweight network that combines global, part-based, and channel features in a unified multi-branch architecture that builds on the resource-efficient OSNet backbone. Using a well-founded combination of training techniques and design choices, our final model achieves state-of-the-art results on CUHK03 labeled, CUHK03 detected, and Market-1501 with 85.1\% mAP / 87.2\% rank1, 82.4\% mAP / 84.9\% rank1, and 91.5\% mAP / 96.3\% rank1, respectively.
\end{abstract}

\section{Introduction}
\label{sec:intro}
\ac{PREID} 
is an important computer vision task for video surveillance applications. Formally, the problem can be stated as follows~\cite{zheng2016person}. Given a probe image $\mathbf{P}$, and a gallery of $M$ images $\mathcal G = \{ \mathbf{G}_i \}_{i=1}^M$, all of which annotated with an associated identity $\mathrm{id}( \mathbf{G}_{i} ) \in \mathbb N$, the goal is to find a similarity measure $\mathrm{sim}\left(\cdot\right)$ such that
\begin{align}
\label{eq:reid}
    i^* = \argmax_{i=1, \dots, M} \mathrm{sim}( \mathbf{P}, \mathbf{G}_i ) && \Rightarrow && \mathrm{id}(\mathbf{P}) = \mathrm{id}(\mathbf{G}_{i^*}).
\end{align}
 While it is no surprise that the success of deep learning and the need for \ac{PREID} as a processing step for person tracking has resulted in numerous approaches, the problem remains challenging, especially when it comes to balancing performance and low complexity of the models.

 Recently, multiple-branch architectures have been proposed in particular~\cite{wang2018learning,chen2020learning,chen2019mixed,chen2019abd,xie2020learning}. These methods allow the network to focus on different person features in individual branches, e.g., on distinct spatial parts or channels. Although branching generally increases model performance, it comes with higher computational costs, especially if the number of branches or the total number of operations in them is increased. We claim that additional model complexity is not necessary and propose a network that outperforms other multi-branch approaches  by using a suitable feature extractor and the right combination of training techniques.

 The resulting network consists of three branches that optimize the global, partial, and channel-wise representations using simple computations, respectively. Despite this branching, we succeed in keeping the number of parameters low using OSNet~\cite{zhou2019omni}, a lightweight feature extractor that has recently proven to be more efficient and accurate than other backbones for \ac{PREID} tasks. Our deep neural network achieves state-of-the-art results on two important benchmark datasets, Market-1501~\cite{zheng2015scalable} and CUHK03 \cite{li2014deepreid}. In detailed ablation studies, we demonstrate how the respective branches increase model performance, why our network performs better than other multi-branch approaches, and what training techniques are necessary to train a multi-branch architecture with OSNet backbone. Code and pretrained models of our research are publicly available\footnote{https://github.com/jixunbo/LightMBN}.

\section{Related Work}
\label{sec:rw}
While \ac{PREID} has been studied as a computer vision task for a long time \cite{gheissari2006person}, deep learning accelerated the research progress and model performance significantly, dominating the scene ever since \cite{zheng2016person, ye2020deep, luo2019bag, zhou2019omni, sun2018beyond, wang2018learning}. \ac{PREID} approaches can be categorized as follows.
First, several methods focus on improving feature extraction for the global input images \cite{luo2019bag, zhou2019omni, dai2019batch}. Luo \etal \cite{luo2019bag} contributed with comprehensive research of many training techniques and were able to find combinations that boost the overall performance. Zhou \etal \cite{zhou2019omni}, on the other hand, concentrated on the feature extraction itself, proposing OSNet, a multi-scale network designed explicitly for the \ac{PREID} task that outperforms standard ResNet50~\cite{he2016deep} backbones despite a much lower number of parameters.

Another important research direction is finding spatial partitions of the persons' images \cite{sun2018beyond, zheng2019pyramidal, chen2020learning}. Usually, the input image is divided into disjoint parts, often horizontal stripes, to obtain partitioned features that are discriminative for person matching. Sun \etal \cite{sun2018beyond} utilized the idea of part pooling, where the partitioning is done via spatial pooling after the convolutional layers of the backbone. This idea has since been used in other architectures \cite{wang2018learning, chen2020learning, zheng2019pyramidal}. In this context, many multi-branch or multi-stage approaches have been developed \cite{wang2018learning, chen2020learning, xie2020learning}. They mostly try to learn global and spatial part features in individual branches or combine part, channel, and global features, either through pooling \cite{wang2018learning, chen2020learning, zheng2019pyramidal} or attention \cite{chen2019mixed, chen2019abd, chen2020salience, lawen2020compact}.

\section{Methodology}
\label{sec:metho}
\begin{figure*}
    \centering
    \includegraphics[width=1.0\linewidth]{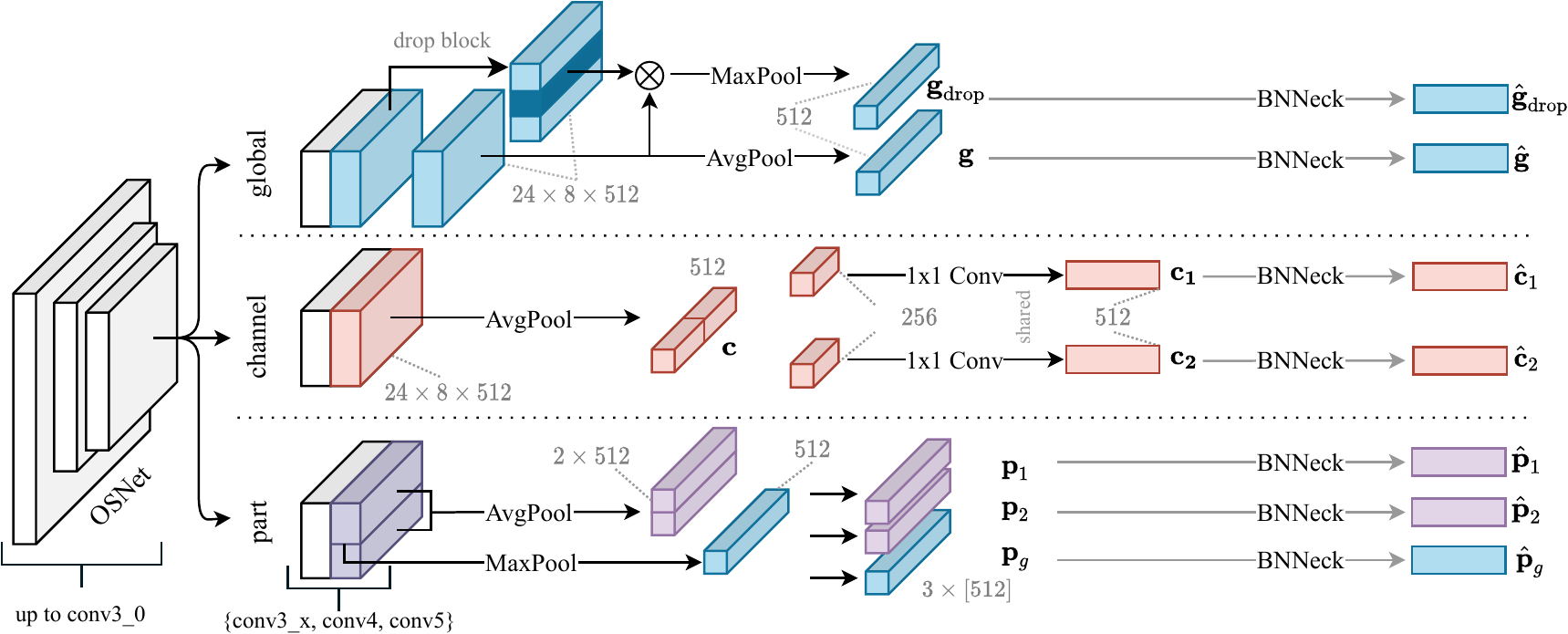}
    \figcaption{Structure of our network.}{After forwarding images through the first three blocks of an OSNet backbone, our network continues in three distinct branches to learn global, channel-based and part-based features. All volumes are forwarded to BNNeck layers to produce final embeddings suited for different loss functions.}
    \label{fig:network}
\end{figure*}

\subsection{Network Architecture}
Like all recent works on the problem, we design an end-to-end neural network architecture based on strong image feature extraction backbones pretrained on ImageNet~\cite{krizhevsky2012imagenet}. In this subsection, we describe the architecture and training of the proposed network to solve Eq.~(\ref{eq:reid}). Our goal is to utilize a multi-branch architecture similar to MGN~\cite{wang2018learning} and SCR~\cite{chen2020learning} that leverages global, part-based and channel-based features, while keeping the overall number of parameters and embeddings low. Consequently and as illustrated in Fig.~\ref{fig:network}, our network consists of three branches: The global branch, the part branch, and the channel branch.

Let $\mathbf{X} \in \mathbb R^{384 \times 128 \times 3}$ be an input image. Before separating into distinct branches, the image $\mathbf X$ is passed through a truncated OSNet~\cite{zhou2019omni} backbone, up until the first layer of the third block, i.e., \emph{conv3\_0}, as in \cite{xie2020learning}. This concept has been employed before with ResNet50 \cite{wang2018learning, chen2020learning}, using the first blocks up to \emph{conv4\_0}. We chose OSNet over ResNet due to its superior performance and lower complexity for \ac{PREID} tasks~\cite{zhou2019omni,xie2020learning}. After forwarding $\mathbf{X}$ through the initial layers, the network forms the three branches, which comprise the remaining layers of OSNet up to the fifth block. By this design, only the layers up to \emph{conv3\_0} are shared by all the branches, and for each individual branch, we obtain a tensor of dimension $24 \times 8 \times 512$.


In the \emph{global branch}, we obtain two global representations as follows: First, we aggregate the information by applying 2D average pooling on the tensor, obtaining the $512$-dimensional vector $\mathbf{g}$. For the second global representation, the initial $24 \times 8 \times 512$-tensor is used as an input for a \emph{drop block}, inspired by \cite{quispe2020top}. The drop block removes the highest activated horizontal regions from the tensor, forcing the network to emphasize on less discriminative regions, which increases the robustness of the resulting representation. Having removed the regions of highest activity, we apply 2D max pooling on the resulting tensor, obtaining another $512$-dimensional vector $\mathbf{g}_{\text{drop}}$.

In the \emph{channel branch}, the initial $24 \times 8 \times 512$-tensor is reduced to a $512$-dimensional vector and then partitioned into two vectors of length $256$ each. We use $1 \times 1$ convolutions to scale the representations back up, obtaining two $512$-dimensional vectors $\mathbf{c}_1$ and $\mathbf{c}_2$. Here, the parameters of the $1 \times 1$ convolutions are shared among both channel parts.

Finally, in the \emph{part branch}, we transform the initial $24 \times 8 \times 512$-tensor into three representations. We use average pooling to obtain a volume of size $2 \times 1 \times 512$ that we split into two $512$-dimensional part-based representations $\mathbf{p}_1$ and $\mathbf{p}_2$, representing the upper and lower body, respectively. Additionally, we use max pooling on the initial volume, obtaining another $512$-dimensional global representation $\mathbf{p}_g$ within the part branch.

We use a BNNeck~\cite{luo2019bag} for all branch vector representations calculated in this way. Each BNNeck block consists of batch normalization and a fully connected to number-of-classes layer. The aim of this block is to optimize embeddings for two different metric spaces at the same time. Embeddings obtained \emph{before} the batch normalization layer are used for optimization with respect to a ranking loss (e.g., triplet loss \cite{schroff2015facenet}), while embeddings obtained \emph{after} the fully connected layer are used for optimization with respect to an identity loss (e.g., \ac{CE} loss). Embeddings obtained \emph{after} the batch normalization but \emph{before} the fully connected layer find a balance between the representations of the two different metric spaces (i.e., ranking space and identitiy space) and are therefore used for inference. From the resulting embeddings we form two sets, given by
\begin{align}
    \mathcal I &:= \left\{ \hat{\mathbf{g}}, \hat{\mathbf{g}}_{\text{drop}}, \hat{\mathbf{p}}_{1}, \hat{\mathbf{p}}_{2}, \hat{\mathbf{p}}_{g}, \hat{\mathbf{c}}_{1}, \hat{\mathbf{c}}_{1} \right\}, \\
    \mathcal R &:= \left\{ \mathbf{g}, \mathbf{g}_{\text{drop}}, \mathbf{p}_{g} \right\},
\end{align}
for training in identity and rank spaces, respectively, where $\hat{\cdot}$ denotes the tensors of the BNNeck representations after the fully connected layer.

\subsection{Training and Loss Functions}
\label{sec:training}
For training, we use a combination of \ac{CE} loss and \ac{MS} loss ~\cite{wang2019multi}. The latter was designed to take advantage of existing pair-wise methods and sampling strategies by exploiting a soft weighting scheme that  considers both self-similarity and relative similarity. We compute MS loss $\mathcal L_{\text{MS}}$ for global embeddings $\mathcal R$ obtained before batch normalization, and CE loss $\mathcal L_{\text{CE}}$ on all embeddings $\mathcal I$ obtained after applying softmax activation to the fully connected layer, i.e.,
\begin{align}
    \mathcal L_{\text{MS}}\left( f(\mathbf{X}), y) \right) &:= \sum_{\mathbf{r} \in \mathcal R} \mathcal L_{\text{MS}} \left( \mathbf r, y \right), \\
    \mathcal L_{\text{CE}}\left( f(\mathbf{X}), y) \right) &:= \sum_{\mathbf{i} \in \mathcal I} \mathcal L_{\text{CE}} \left( \mathbf i, y \right),
\end{align}
where $f(\mathbf X)$ is our networks output when forwarding $\mathbf X$. For \ac{CE} loss $\mathcal L_{\text{CE}}$, we further use \emph{label smoothing}~\cite{szegedy2016rethinking, luo2019bag}, which is a regularization technique that encourages the model not to be too confident on the training data. It adds a uniform noise distribution in \ac{CE} calculation to soften the ground truth labels, which helps to improve model generalization.
Thus, the overall objective loss function is
\begin{equation}
\mathcal{L}= \lambda_{\text{CE}} \mathcal{L}_{\text{CE}} + \lambda_{\text{MS}} \mathcal{L}_{\text{MS}},
\end{equation}
where $\lambda_{\text{CE}}$ and $\lambda_{\text{MS}}$ are suitable weights.
Additionally, we use \emph{random erasing augmentation} (REA)~\cite{zhong2017random}, which randomly substitutes a rectangle with the image's mean value. It has demonstrated to improve  model generalization and to produce higher variance training data.
Cosine annealing strategies are common in \ac{PREID} networks \cite{zhou2019omni, zhu2020voc}. To further boost performance, we use warm-up cosine annealing \cite{goyal2017accurate, he2019bag} as our learning rate strategy rather than traditional step learning rate schedules. The learning rate first grows linearly from $6 \cdot 10^{-5}$ to $6 \cdot 10^{-4}$ in 10 epochs, then cosine decay to $6 \cdot 10^{-7}$ is applied in the remaining epochs. The learning rate $\operatorname{lr}(t)$ at epoch $t$ with $T$ total epochs is given by
$$
\operatorname{lr}(t)=\left\{\begin{array}{ll}
6 \cdot 10^{-4} \cdot \frac{t}{10}, & \text { if } t \leq 10 \\
6 \cdot 10^{-4} \cdot \frac{1}{2} \left( 1 + \cos\left(\pi \frac{t-10}{T -10}\right) \right), &  \text { if } 10<t \leq T.
\end{array}\right.
$$

\section{Experimental Results}
\label{sec:exp}

\begin{table*}[!htbp]
    \figcaption{Comparison of our method with state-of-the-art.}{The table lists our results on the two most used benchmarks, Market-1501 and CHUK03. The latter was evaluated on the labeled set (CHUK03-L) and the detection set (CHUK03-D) in multi-gallery-shot setting (cf. \cite{zhong2017re}). Note that all results are reported \emph{without} re-ranking (cf. \cite{zhong2017re}).}
\label{tab:stoa}

\centering
\setlength{\tabcolsep}{10pt}
\small
\resizebox{\textwidth}{!}{
\begin{tabular}{llllrrrrrrrr}
\toprule
 &    &   &   & \multicolumn{2}{c}{Market-1501}  & \multicolumn{2}{c}{CHUK03-L}	& \multicolumn{2}{c}{CHUK03-D} \\
\cmidrule(lr){5-6} \cmidrule(lr){7-8} \cmidrule(lr){9-10} \cmidrule(lr){11-12}
Type&Method &Publication  &Backbone	   & r1 	& mAP	&r1 	& mAP  &r1 	& mAP 	 \\
\midrule
\multirow{3}{*}{Global feature}&BagOfTricks~\cite{luo2019bag}&CVPRW'19 & ResNet50   &94.5	&85.9	&-- 	&-- &-- &--		\\
&OSNet~\cite{zhou2019omni} &ICCV'19  &OSNet    &94.8   &84.9 & -- & -- &72.3 &67.8  \\
&BDB~\cite{dai2019batch}&ICCV'19  &ResNet50        & 95.3 &86.7	&79.4	&76.7 & 76.4 & 73.5		\\
\midrule
\multirow{3}{*}{Part-based}&PCB+RPP~\cite{sun2018beyond}&ECCV'18 &ResNet50   & 93.8	&81.6	&--	&-- &-- & --		\\
&MGN~\cite{wang2018learning}&ACM MM 18   &ResNet50    &95.7 &  86.9 &68.0 & 67.4 & 66.8 & 66.0    \\
&Pyramid~\cite{zheng2019pyramidal}&CVPR'19 & ResNet101  & 95.7 &88.2	&78.9	&76.9 &79.9 & 74.8		\\
&SCR~\cite{chen2020learning}&WACV'20 & ResNet50 & 95.7& 89.0 & 83.8 & 80.4 &  82.2 & 77.6\\
\midrule
\multirow{3}{*}{Attention-based}&MHN~\cite{chen2019mixed}&ICCV'19 &ResNet50        & 95.1  &  85.0 &  77.2 & 72.4 & 71.7 & 65.4\\
&ABD~\cite{chen2019abd}&ICCV'19    &ResNet50        & 95.6  &  88.3 &   -- & -- & -- & --\\
&PLR-OSNet~\cite{xie2020learning}& PRCV '20&OSNet    & 95.6 & 88.9 & 84.6 & 80.5 & 80.4 & 77.2 \\
&SCSN~\cite{chen2020salience}&CVPR'20                  &ResNet50        & 95.7  &  88.5 & 86.8 & 84.0  & 84.7 & 81.0\\
&Compact Re-ID \cite{lawen2020compact} & ACM ICMR '20 & other & 96.2 &	89.7 & -- & -- & -- & -- \\
    \midrule
 \textbf{Ours} &\textbf{LightMBN}  &  & OSNet   & \textbf{96.3} & \textbf{91.5} & \textbf{87.2} & \textbf{85.1} & \textbf{84.9} & \textbf{82.4} \\
 &LightMBN  (computed via \cite{zheng2015scalable}) &  & OSNet   & \textbf{96.3} & 91.2 & \textbf{87.2} & 83.8 & \textbf{84.9} & 81.0 \\

    \bottomrule

    \end{tabular}}
\end{table*}

\nparagraph{Datasets} We evaluated the model on two of the most widely used large-scale datasets, Market-1501~\cite{zheng2015scalable} and CUHK03 \cite{li2014deepreid}. The Market-1501 dataset contains 32,668 images of 1,501 persons across 6 cameras, whereas the CUHK03 dataset comprises 13,164 images of 1,360 person across 6 cameras. For CUHK03, we use the new 767-split protocol~\cite{zhong2017re}, obtaining results for the labeled (CUHK03-L) and detected (CUHK03-D) configurations separately. We did not evaluate on DukeMTMC-ReID since use of this dataset has been prohibited by the authors.

\nparagraph{Training Details} For training, input images are normalized to channel-wise zero-mean and a standard variation of 1 and spatial resolution of $384 \times 128$. Data augmentation is performed by resizing images to 105\% width and height and random cropping, as well as random horizontal flip with a probability of 0.5. Models are trained for 140 epochs for Market-1501 and 180 epochs for CUHK03 with a batchsize of 48. A batch consists of 8 samples for 6 identities each. The parameters are optimized by using using the Adam optimizer\,\cite{kingma2014adam} with $\epsilon = 1e-8 $, $\beta_1 = 0.9$ and $\beta_2 = 0.999$. The backbones are pre-trained on ImageNet~\cite{krizhevsky2012imagenet} and all experiments are implemented with PyTorch\,\cite{paszke2017automatic}. To balance the losses we chose $\lambda_{\text{CE}} = \lambda_{\text{MS}} = 0.5$.

\nparagraph{Evaluation Details} Cosine distance is utilized to compute cumulative matching characteristics\,(CMC)~\cite{moon2001computational}. Query and gallery images are re-sized to $384 \times 128$ pixels and normalized. For a fair comparison with other existing methods, the CMC rank-1 accuracy (r1) and mean Average Precision\,(mAP) are reported as evaluation metrics. Results with the same identity and the same camera ID as the query image are not counted. The authors of \cite{hermans2017defense} state in their official code repository\footnote{https://github.com/VisualComputingInstitute/triplet-reid} that mAP values computed with recent \ac{PREID} frameworks are about 1\%-point higher than those computed by the original Matlab evaluation code of Market-1501 \cite{zheng2015scalable}. We were able to reproduce this. For completeness and fair comparison, we also state the mAP values for our final models as computed by the original evaluation script. 
We hope to raise more awareness to this issue by providing both results.

\begin{table}[htbp]
\figcaption{Ablation study of branch influences.}{We investigate our models performance under the specified branch configurations, where G+C+P refers to our original model.}
\label{tab:ablation1}

\centering
\setlength{\tabcolsep}{10pt}
\renewcommand\arraystretch{1.1}
\small

\begin{tabular}{lrrrrr}
\toprule

                                         & \multicolumn{2}{c}{Market-1501} & \multicolumn{2}{c}{CUHK03-D} \\

                                         \cmidrule(lr){2-3} \cmidrule(lr){4-5}
Branch                                   & rank1            & mAP          & rank1             & mAP           \\
\midrule
 Global (G)                                 & 95.4            & 89.3           &       80.8      &    77.3     \\
 Channel (C)                                & 95.9            & 88.8           &       74.7      &    71.2        \\
 Part (P)                                   & 95.9          &   90.2         &       80.3      &    77.9 \\
\midrule
 C+P                                     &  96.1         & 91.2          &  82.7      & 79.8      \\
 G+C                                     &  96.0          & 90.9          &  82.0      & 79.7            \\
 G+P                                     &    96.1         & 91.2          &  83.4      & 81.3             \\
 G+C+P                                   &    96.3        & 91.5           &  84.9      &  82.4           \\
\bottomrule
\end{tabular}
\end{table}

\subsection{Comparison with State-of-the-Arts}
Table \ref{tab:stoa} compares the performance of our model with that of other recent methods. Our model achieves state-of-the-art results on Market-1501, CUHK-L and CUHK-D, both in terms of rank-1 accuracy and mAP. The large difference in performance with regard to the mAP on all datasets is particularly noticeable. Interestingly, despite its simplicity, our architecture achieves better performance than other multi-branch approaches. Architecturally, our model is closely related to previous work such as MGN \cite{wang2018learning}, PLR-OSNet \cite{xie2020learning}, and, in particular, SCR \cite{chen2020learning}. All of these approaches use a truncated backbone followed by branching. MGN relies on ResNet50 and only uses spatial partitions, whereas our model builds upon OSNet and also better exploits the \ac{PREID} problem by additionally using channel partitions. In this regard, SCR is the most similar architecture since both spatial and channel partitions are used for multi-loss training. However, for good performance, SCR requires nearly twice as many embeddings as our model and creates part and channel partitions in the same branch, which could theoretically impede the branches' specialization.

\subsection{Ablation Study}
\nparagraph{Influence of Branches}
When introducing branches to a neural network architecture, the parameter count can raise substantially. Thus, any such introduction has to be well-justified. Table~\ref{tab:ablation1} depicts our network's performance for different branch combinations. The results suggest that single branches perform similarly when the other two respective branches are deactivated. Among all branches, the channel branch has the lowest performance on CUHK03-D, indicating that global features are very important for generalization on this dataset. As can be seen by the pairwise combination of branches, the part branch influences the performance significantly on CUHK03-D. By using all three branches together, our model achieves state-of-the-art results on both datasets.

\nparagraph{Influence of Backbones} Table~\ref{tab:ablation2} shows some examples of the different performances of ResNet50 and OSNet. The raw model with ResNet50 (i.e., the one without beneficial additions) has the weakest performance among all models. Only with all possible additions it is able to achieve  similar performance of a raw model with OSNet backbone. The best configuration that can be achieved with ResNet50 is still inferior than our final model. Our model with OSNet backbone only has about 9 million parameters, compared to about 23 million with ResNet50 backbone.

\nparagraph{Influence of Learning Rate Schedule}
As can be seen in Table~\ref{tab:ablation2}, when substituting the cosine warmup annealing schedule with a constant schedule, performance decreases. For the constant schedule, we have reduced the initial learning rate of $6 \times 10^{-4}$ three times by a factor of $10$ in the 50th, 80th and 110th epoch, respectively. The results indicate the importance of a suitable learning rate strategy for PRID on both datasets.

\nparagraph{Influence of Drop Block} The results in Table \ref{tab:ablation2} suggest that the drop block has hardly any influence on the performance on Market-1501. On the other hand, results on the CUHK03 dataset clearly show that the drop block can lead to better generalization on the test set and increases both metrics.

\begin{table}[htbp]
\figcaption{Ablation study of training techniques.}{We investigate our models performance under the specified training modifications. Here, WCA indicates use of warmup cosine annealing, MS the use of \ac{MS} loss over triplet loss, DB the use of drop block, and OSNet the use of OSNet over ResNet50 as backbone, respectively.}
\label{tab:ablation2}

\centering
\setlength{\tabcolsep}{10pt}
\renewcommand\arraystretch{1.1}
\small

\resizebox{\linewidth}{!}{
\begin{tabular}{ccccrrrrr}
\toprule

 \multicolumn{4}{c}{Configuration}                                       & \multicolumn{2}{c}{Market-1501} & \multicolumn{2}{c}{CUHK03-D}  \\

                                                   \cmidrule(lr){1-4}                \cmidrule(lr){5-6}  \cmidrule(lr){7-8}
  OSNet & WCA  & MS & DB                               & r1            & mAP          & r1             & mAP \\
\midrule
  \xmark& \xmark &  \xmark & \xmark &     95.4      & 87.9       & 71.7           & 70.3\\
    \xmark& \cmark & \cmark  & \cmark   & 96.1      & 90.4       & 81.0           & 79.1 \\
  \cmark & \xmark &  \xmark & \xmark &    96.1      & 90.2 & 78.2 & 75.2 \\
  \cmark & \cmark &  \xmark & \xmark &  96.2 & 91.1              & 83.5 & 81.1\\
  \cmark & \cmark & \xmark  & \cmark   &  96.3     &   91.5     & 83.2           & 80.9\\
  \cmark & \xmark &  \cmark  & \cmark    & 96.0      & 90.6       & 78.8           & 76.3\\
   \cmark & \cmark & \cmark  & \xmark    & 96.2      &   91.2         &  83.4  &	80.9\\
  \cmark & \cmark &  \cmark  & \cmark    & 96.3      & 91.5       & 84.9           & 82.4 \\

\bottomrule
\end{tabular}}
\end{table}

\nparagraph{Influence of Loss Functions} We trained various modifications of our model with triplet loss instead of MS loss. Using MS loss in the final model slightly increases the rank-1 and mAP performance on CUHK03, but not on Market-1501. Thus, the choice of ranking loss function can be important for generalization on smaller datasets.

\section{Conclusion}
\label{sec:conclusion}
We have presented a multi-branch neural network that achieves state-of-the-art results on Market-1501 and CUHK03. Although branches increase the overall parameter count, we can keep the overall model complexity low by utilizing a lightweight OSNet backbone and suitable training techniques. The distinct branches of our network can capture the essential person features. Overall our research suggests that learning rate schedules and the backbone choice heavily influence the model performance and that drop blocks and MS loss assist the model in generalizing the smaller CUHK03 dataset. We conclude that multi-branch architectures should focus on the right combination of training techniques and OSNet feature extraction in favor of adding model complexity.

\bibliographystyle{IEEEbib}
\bibliography{strings,refs}

\end{document}